\documentclass[nohyperref]{article}
\usepackage{lmodern}
\usepackage{microtype}
\usepackage{graphicx}
\usepackage{subfigure}
\usepackage{booktabs} %
\usepackage{hyperref}

\usepackage[accepted]{icml2022_arxiv}
\usepackage{url}
\usepackage{amsmath}
\usepackage{amssymb}
\usepackage{mathtools}
\usepackage{amsthm}
\usepackage{mathtools}
\usepackage{placeins}
\usepackage{enumerate}
\usepackage{enumitem}
\usepackage{float}
\usepackage[usenames,dvipsnames]{xcolor}

\usepackage{thmtools}
\usepackage{thm-restate}
\usepackage{balance}

\newcommand{\defeq}{\vcentcolon=}
\newcommand{\reals}{\mathbb{R}}
\newcommand{\pistar}{\pi^\star}
\newcommand{\qstar}{Q^\star}

\newcommand{\states}{\mathcal{S}}
\newcommand{\actions}{\mathcal{A}}
\newcommand{\goals}{\mathcal{G}}
\newcommand{\rewards}{\mathcal{R}}
\newcommand{\transition}{\mathbb{P}}
\newcommand{\state}{\mathbf{s}}
\newcommand{\action}{\mathbf{a}}
\newcommand{\entity}{e}
\newcommand{\entities}{\mathcal{E}}
\newcommand{\subgoals}{\mathfrak{g}}
\newcommand{\goal}{\mathbf{g}}
\newcommand{\subgoal}{g}
\newcommand{\agent}{u}
\newcommand{\agents}{\mathcal{U}}
\newcommand{\subreward}{\tilde{r}}

\newcommand{\flatten}{\text{vec}}

\newcommand{\fnsnp}[1]{\textit{{#1}-Switch~+~{#1}-Push}}

\hypersetup{
 colorlinks=True,
 linkcolor=RoyalBlue,
 citecolor=RoyalBlue,
 urlcolor=RoyalBlue}

\newtheorem{definition}{Definition}

\newtheorem{property}{Property}

\usepackage[textsize=tiny]{todonotes}

\icmltitlerunning{Policy Architectures for Compositional Generalization in Control}

\begin{document}

\twocolumn[
\icmltitle{Policy Architectures for Compositional Generalization in Control}

\begin{icmlauthorlist}
    \icmlauthor{Allan Zhou}{Stanford}
    \icmlauthor{Vikash Kumar}{FAIR}
    \icmlauthor{Chelsea Finn}{Stanford}
    \icmlauthor{Aravind Rajeswaran}{FAIR}
\end{icmlauthorlist}

\icmlaffiliation{Stanford}{Stanford University}
\icmlaffiliation{FAIR}{Meta AI (FAIR)}

\icmlcorrespondingauthor{Allan Zhou}{\texttt{ayz@stanford.edu}}
\icmlcorrespondingauthor{Aravind Rajeswaran}{\texttt{aravraj@fb.com}}

\icmlkeywords{Reinforcement Learning, Compositionality, Self-Attention}

\vskip 0.3in
]

\printAffiliationsAndNotice{}  %

\begin{abstract}
Many tasks in control, robotics, and planning can be specified using desired goal configurations for various entities in the environment. Learning goal-conditioned policies is a natural paradigm to solve such tasks. However, current approaches struggle to learn and generalize as task complexity increases, such as variations in number of environment entities or compositions of goals.
In this work, we introduce a framework for modeling entity-based compositional structure in tasks, and create suitable policy designs that can leverage this structure.
Our policies, which utilize architectures like Deep Sets and Self Attention, are flexible and can be trained end-to-end without requiring any action primitives. 
When trained using standard reinforcement and imitation learning methods on a suite of simulated robot manipulation tasks, we find that these architectures achieve significantly higher success rates with less data. We also find these architectures enable broader and compositional generalization, producing policies that \textbf{extrapolate} to different numbers of entities than seen in training, and \textbf{stitch} together (i.e. compose) learned skills in novel ways. Videos of the results can be found at \url{https://sites.google.com/view/comp-gen-rl}.
\end{abstract}

\section{Introduction}
\label{sec:intro}
\begin{figure}[t!]
    \centering
    \includegraphics[width=\columnwidth]{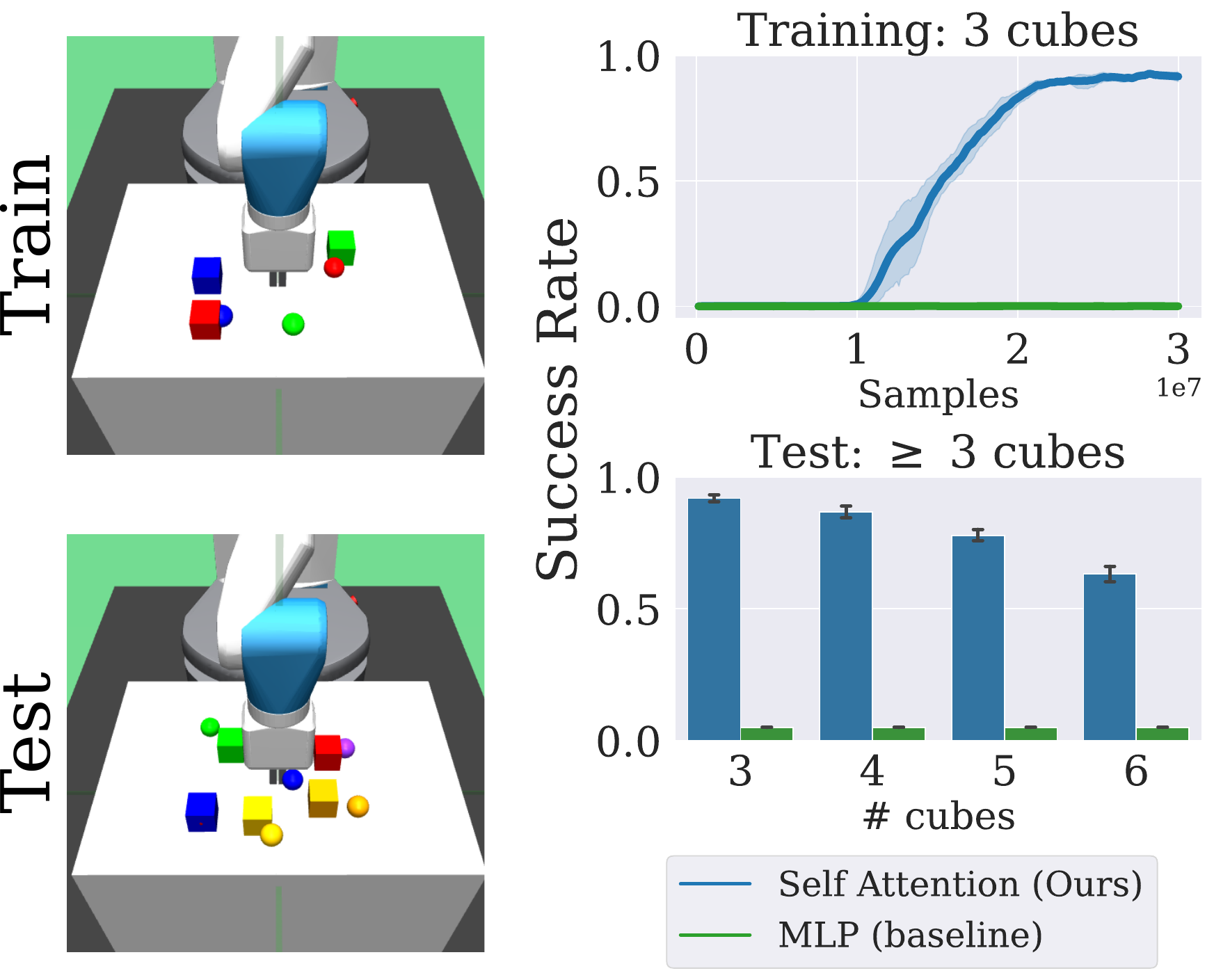}
    \vspace*{-1em}
    \caption{
    A family of tasks where the agent is trained to re-arrange three cubes (top-left), but tested zero-shot to re-arrange anywhere from three to six cubes (bottom-left). Reinforcement learning (RL) with standard MLPs fails to even learn the 3-cube task. Our self-attention based policy can learn the three cube task, and also solve tasks with more cubes using no additional data.}
    \vspace*{-15pt}
    \label{fig:teaser_figure}
\end{figure}

Goal specification is a powerful abstraction for training and deploying AI agents~\citep{KaelblingGoals,schaul2015universal,her}. For instance, object reconfiguration tasks~\citep{batra2020rearrangement}, like loading plates in a dishwasher or arranging pieces on a chess board, can be described through spatial (6DOF pose) and semantic (on vs off) goals for various objects. Furthermore, a broad goal for a scene can be naturally described through compositions of goals for individual entities. In this work, we introduce a framework for modeling tasks with this \textit{entity-centric} compositional structure, and study policy architectures that can utilize such structural properties. Our framework and designs are broadly applicable for goal-conditioned reinforcement and imitation learning. Through experiments in a suite of simulated robot manipulation environments, we find that our policy architectures learn substantially faster compared to standard multi-layer perceptrons (MLPs) and demonstrate significantly improved generalization capabilities, a preview of which is depicted in Figure~\ref{fig:teaser_figure}.

Consider the motivating task of arranging pieces on a chess board using a robot arm. A naive specification would provide goal locations for all 32 pieces simultaneously. However, we can immediately recognize that the task is a composition of 32 sub-goals involving the rearrangement of individual pieces. This understanding of compositional structure can allow us to focus on one object at a time, dramatically reducing the size of effective state space and help combat the curse of dimensionality that plagues RL~\citep{SuttonBook, BertsekasBook}. Moreover, such a compositional understanding would make an agent invariant to the number of objects, enabling generalization to fewer or more objects. Most importantly, it can enable reusing shared skills like pick-and-place, enhancing learning efficiency. Finally, we also note that a successful policy cannot completely decouple the sub-tasks and must consider their interactions. For example, if a piece must be moved to a square currently occupied by another piece, the piece in the destination square must be moved first. 

A straightforward policy architecture like the commonly used MLP does not have the inductive biases or structure to exhibit the aforementioned compositional properties. To overcome these deficiencies, we turn to the general field of ``geometric deep learning''~\citep{GeoDL} which is concerned with the study of structures, symmetries, and invariances exhibited by function classes. The main contributions of our work include the development of the Entity-Factored MDP (EFMDP) framework which clearly exposes the structures and symmetries of many compositional tasks. We develop new policy and critic architectures for deep reinforcement and imitation learning that leverage the structural properties of EFMDPs, drawing upon advances in geometric deep learning like Deep Sets~\citep{deepsets} and Attention~\cite{transformer}. Our architectures are flexible and broadly applicable, do not assume any low-level primitives or options, and can be trained end-to-end using standard RL and IL algorithms. Through experimental evaluations on a suite of simulated robot manipulation tasks, some of which are illustrated in Figure~\ref{fig:task_teaser}, we find that compared to standard MLPs our architectures learn complex tasks more than $4\times$ faster and achieve greater than $15\times$ higher success rates when generalizing to new tasks.

\section{Related Work}

\textbf{Compositionality and Hierarchy.} One approach to solving long-horizon or sequential tasks is to explicitly learn ``macro'' actions that correspond to useful skills, which can then be sequenced by a high-level policy. This approach is commonly taken in hierarchical RL and related frameworks \citep{FeudalRl,ParrHierarchiesRl,HierarchicalMaxQ,OptionCritic,hiro,LanguageHierarchicalRl, lin2022efficient}. Our approach also enables the learning of long-horizon and compositional tasks, but simply through architectural modifications to the policy in end-to-end learning, as opposed to explicitly learning action representations or modifying the training process.

\textbf{Entity-centric modeling.}  
We introduce and study a class of learning problems where an agent must learn in an environment containing a number of entities. Prior works have explored different ways of modeling relationships between entities in the environment, for instance through use of graph neural networks~\citep{NeuralTaskGraphs,bapst2019structured,li2020multiobject,veerapaneni2020entity,lin2022efficient}, transformers~\citep{relationalrl,carvalhoroma,tang2021sensory}, or other means~\citep{goyal2019recurrent,deepsetsrl}. We introduce a framework for entity-based compositionality in reinforcement learning and study both theoretically and empirically the structural invariances in these classes of learning problems, and how various architectures may utilize these structural properties.

\textbf{Policy Architectures in RL.} 
Compared to other areas of deep learning, RL has seen limited architectural innovations, with small MLPs and LSTMs still being the dominant architectures in continuous control~\citep{ddpg,ppo,sac}. More recently, a number of works have started exploring deeper and sequence modeling architectures for RL. \citet{D2RL} study deeper networks for continuous control with DenseNet-style~\citep{DenseNet} connections. \citet{DecisionTransformer} and \citet{TrajectoryTransformer} explored the use of transformers for decision making and control. While our work also studies the use of self-attention for control, we do not use it to model the trajectory history. Instead, we use self-attention for modeling the interactions between entities and goals to enable compositional generalization. Thus our work is complementary to advances in \cite{DecisionTransformer}.

\section{Problem Formulation and Architectures}

In this section, we first formalize our problem setup by introducing the entity-factored MDP. This setting is capable of modeling many applications including table-top manipulation, scene reconfiguration, and muti-agent learning. Subsequently, we also introduce policy architectures that can enable efficient learning and generalization by utilizing the unique structural properties of the entity-factored MDP. 

\subsection{Problem Setup}

We study a learning paradigm where the agent can interact with many entities in an environment. The task for the agent is specified in the form of goals for some subset of entities (including the agent). We formalize this learning setup with the Entity-Factored Markov Decision Process (EFMDP).

\begin{figure}[t!]
    \centering
    \includegraphics[width=0.7\linewidth]{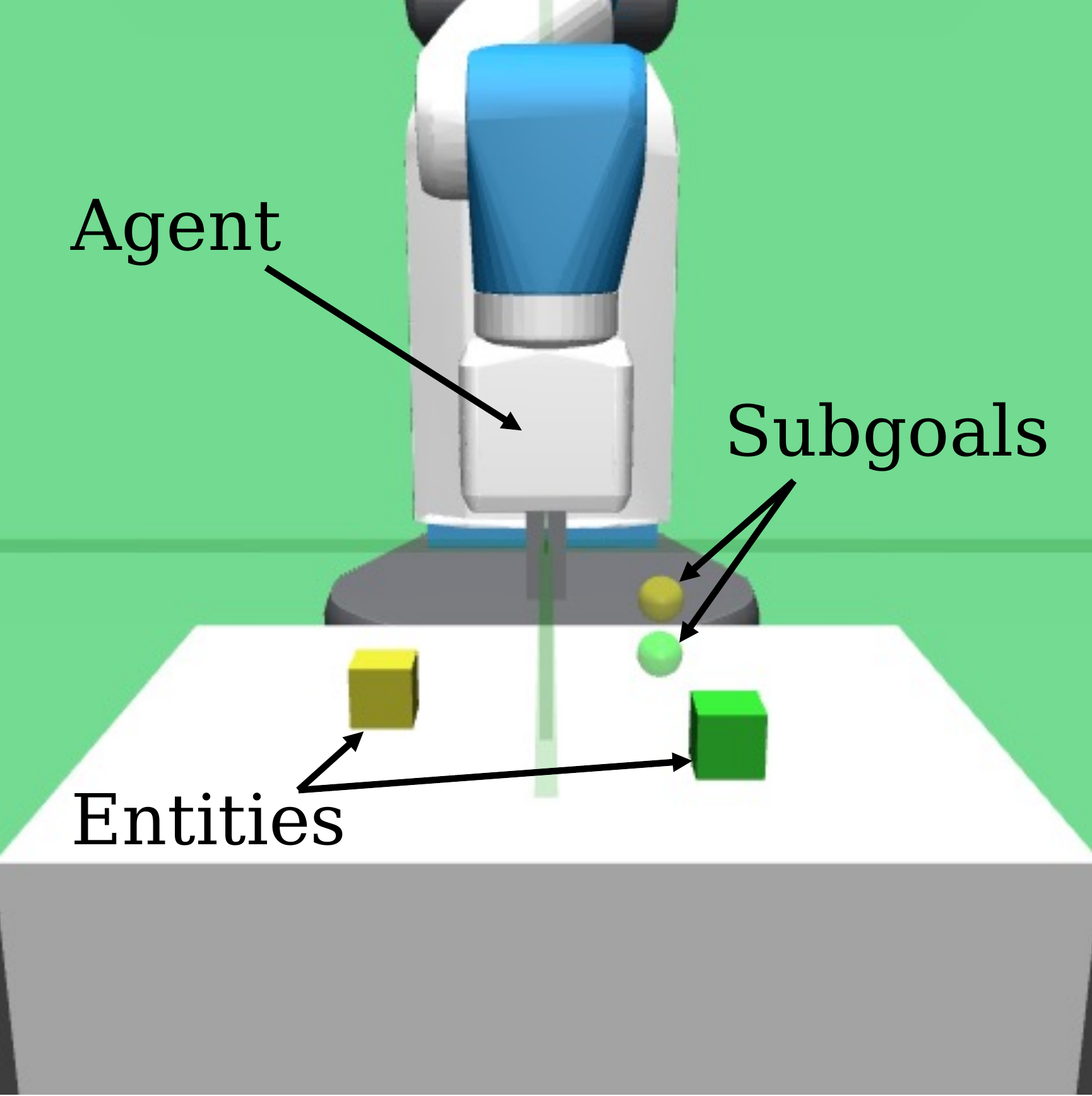}
    \caption{In an EFMDP, an agent interacts with entities that have corresponding subgoals. This framework can model rearrangement, strategic game playing, and multi-agent systems. In this ``pushing and stacking'' example, the agent must move the green cube to its subgoal (indicated by the green sphere), then stack the yellow cube on top of the green cube.}
    \label{fig:cartoon}
\end{figure}

\begin{definition}[Entity-Factored MDP]
\label{def:mdp}
An EFMDP with $N$ entities is described through the tuple: $\mathcal{M}^E := \left< \agents, \entities, \subgoals, \actions,  \transition,  \rewards, \gamma \right>$. Here $\agents$ and $\entities$ are the agent and entity state spaces, $\subgoals$ is the subgoal space and $\actions$ is the agent's action space. The overall state space $\states\defeq\agents\times\entities^N$ has elements $\state = (u, \entity_1, \cdots, \entity_N)$ and the overall goal space $\goals\defeq\subgoals^N$ has elements $\goal = (\subgoal_1, \ldots, \subgoal_N)$. The reward and dynamics are described by:
\begin{align}
\rewards\left(\state, \goal \right) &:= \rewards\left( \lbrace \subreward(\entity_i, \subgoal_i, \agent) \rbrace_{i=1}^N \right)\\
\transition(\state'|\state,\action) &\defeq \transition\left(\left(\agent', \{\entity_i'\}_{i=1}^N\right)| \left(\agent, \{\entity_i\}_{i=1}^N\right), \action\right)
\end{align}
for $\state,\state'\in\states$, $\action\in\actions$, and $\goal\in\goals$.
\end{definition}

The EFMDP is a goal-conditioned MDP~\cite{KaelblingGoals,schaul2015universal} with additional structure. Each entity can be associated with a specific reward $\subreward_i = \subreward(\entity_i, \subgoal_i, \agent)$, which are all aggregated together to reward the agent. The aggregation can follow various compositional rules, such as requiring ``all'' entity subgoals to be satisfied or ``any'' entity subgoal be satisfied. We also note that the EFMDP does not force the entities to be exchangeable or indistinguishable, since the entity state space may contain identifying properties distinguishing each entity. The ultimate objective for the learning agent in is to learn a policy $\pistar: \states\times\goals \rightarrow \actions$ that maximizes the long term rewards, given by:
\begin{equation}
\label{eq:objective}
\pistar \defeq \arg\max_{\pi}  \ \ \left\{ J(\pi) := \mathbb{E}_{\pi} \left[ \sum_{t=0}^\infty \gamma^t \rewards(\state_t, \goal) \right]\right\},
\end{equation}
where the expectation is under EFMDP dynamics.

A wide variety of tasks involve an agent interacting with different entities in the environment, and can be cast as EFMDPs. Examples include robotic manipulation, scene reconfiguration, multi-agent learning, and strategic game playing. At the same time, the EFMDP contains more structure and symmetry compared to the standard MDP model, which can enable more efficient learning and better generalization. The crucial symmetry exists in the reward and dynamics, which treat entity-subgoal pairs as unordered sets and are therefore invariant under permutations:
\begin{property}[EFMDP Permutation Symmetry]
\label{property:inv}
For any permutation $\sigma \in S_N$ (the group of all permutations of $N$ items), the reward satisfies $\rewards(\sigma \state, \sigma \goal)=\rewards(\state,\goal)$ and the transition dynamics satisfy $\transition(\sigma \state'|\sigma \state, \action)=\transition(\state'|\state,\action)$ for any $\state,\state'\in\states$ and $\action\in\actions$, where:
\begin{align}
\label{eq:sigma-s}
\sigma\state&\defeq(\agent, \entity_{\sigma(1)}, \cdots, \entity_{\sigma(N)})\\
\label{eq:sigma-g}
\sigma\goal&\defeq(\subgoal_{\sigma(1)}, \cdots, \subgoal_{\sigma(N)})
\end{align}
\end{property}
This property captures the general intuition that the ordering of entity-subgoal pairs is arbitrary and not relevant to the actual environment. We also prove that any optimal policy and the optimal value function are permutation invariant.
\begin{restatable}[Policy and Value Invariance]{proposition}{inv}
\label{prop:inv}
In any EFMDP with $N$ entities, any optimal policy $\pistar:\states\times\goals\rightarrow\actions$ and optimal action-value function $\qstar:\states\times\actions\times\goals\rightarrow\reals$ are both invariant to permutations of the entity-subgoal pairs. That is, for any $\sigma\in S_N$:
\begin{align*}
\pistar(\sigma \state,\sigma\goal) &= \pistar(\state, \goal)\\
\qstar(\sigma \state,\action, \sigma\goal) &= \qstar(\state, \action, \goal)
\end{align*}
\end{restatable}
This is a direct consequence of the permutation symmetry in reward and dynamics; we provide a proof in Appendix~\ref{sec:permutation-invariance}.

In the next subsections, we will use this invariance to guide architecture design in reinforcement and imitation learning on EFMDPs. It turns out that certain ``entity-centric'' designs can achieve policy and value invariance, enabling them to utilize the symmetries and structure of the EFMDP. Although our descriptions will focus on the policy, the core principles (e.g., Proposition~\ref{prop:inv}) will apply similarly to both policies and critics.

\begin{figure*}
    \centering
    \includegraphics[width=\textwidth]{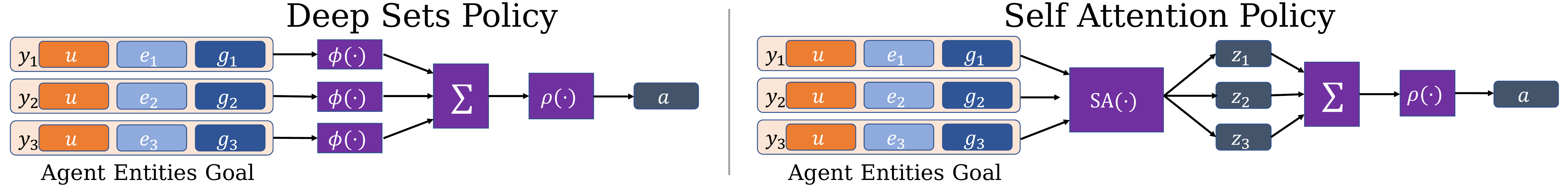}
    \vspace{-1em}
    \caption{Visualizations of implementing an entity-based goal conditioned policy using either Deep Sets (left) or Self Attention (right). The policy $\pi: (\state, \goal)\mapsto \action$ receives state $\state=(\agent, \entity_1,\cdots,\entity_N)$ containing agent state $\agent$ and entity states $\entity_i$. The goal $(\subgoal_1,\cdots,\subgoal_N)$ contains subgoals for each entity. Both policies arrange the input into $N$ vectors $y_i = (\agent, \entity_i, \subgoal_i)$, one per entity. The Deep Set policy processes each $y_i$ independently with MLP $\phi(\cdot)$, aggregates the outputs, and maps the result to an action using MLP $\rho(\cdot)$. The self attention encoder $\text{SA}(\cdot)$ produces output $z_1, \cdots, z_N$ and uses self-attention (Eq.~\ref{eq:self-attention}) to model interactions between the entities/subgoals. The $z_i$ are mapped to an action by summation and an MLP $\rho(\cdot)$.}
    \label{fig:architectures}
\end{figure*}

\subsection{Multilayer Perceptrons (MLPs)}
\label{sec:mlp}
Standard RL and IL approaches assume they are solving a generic MDP, and do not use any additional structure. The generic approach is thus to parameterize the learned policy by an MLP, which takes a fixed size input vector and applies alternating layers of affine transforms and pointwise nonlinearities to produce a fixed size output vector. To implement $\pi(\state, \goal)$ with an MLP we arrange the contents of $(\state, \goal)$ into a single long vector using concatenation:
\begin{equation}
    \label{eq:mlp-input}
    \flatten(\state, \goal)\defeq\text{Concatenate}(\underbrace{\agent, \entity_1,\cdots,\entity_N}_{=\state},\underbrace{\subgoal_1,\cdots,\subgoal_N}_{=\goal})
\end{equation}
Denoting the action of the MLP as a vector-to-vector function $\text{MLP}(\cdot)$, our policy is defined:
\begin{equation}
    \pi(\state, \goal) \defeq \text{MLP}(\flatten(\state, \goal))
\end{equation}
Since MLPs expect input vectors of a fixed dimension, testing on tasks with more entities requires zero padding during training. For example, suppose we are training in an environment with $N$ entities and testing in an environment with $M>N$ entities. During training, we create fake entities and subgoals $(e_{N+1},\cdots,e_M)=(0, \cdots, 0)$ and $(g_{N+1},\cdots,g_M)=(0,\cdots,0)$ that are included in the concatenation of Eq.~\ref{eq:mlp-input}, so that the MLP's input dimension always matches that of the test environment.

\subsection{Deep Sets}
Recall that the optimal policy in an EFMLP should be permutation invariant (Prop.~\ref{prop:inv}). The MLP policy represents a ``black-box'' approach to generic MDPs, and unsurprisingly might require substantial data to learn a permutation invariant policy. The Deep Sets~\citep{deepsets} architecture is better suited to achieve the desired permutation invariance of subgoal-entity pairs. Given a set of vectors $x=\{x_1, \cdots, x_N\}$, it constructs a model of the form:
\begin{equation}
\label{eq:deep-sets}
    \text{DS}(x) \defeq \rho\left(\sum_{i} \phi(x_i)\right),
\end{equation}
where $\rho$ and $\phi$ are themselves typically MLPs. Not only can one easily verify that $\text{DS}(\cdot)$ is permutation invariant, \citet{deepsets} further showed that it is in some sense a universal approximator to \textit{any} permutation invariant function of $x$, given that $\rho,\phi$ are sufficiently expressive. This means that at least in theory, we can represent our permutation invariant policy $\pistar$ using Deep Sets.

Recall that we desire a policy invariant to the ordering of entities and subgoals $\{(\entity_1, \subgoal_1),\cdots,(\entity, \subgoal_N)\}$, but we also need to process the agent state $\agent$. Our approach is to simply share $\agent$ across the set elements:
\begin{align}
    \pi(\state, \goal) &\defeq \text{DS}\left(\{y_i\}_{i=1}^N\right)\\
    \label{eq:deepset-input}
    y_i &\defeq \text{Concatenate}(\underbrace{\agent, \entity_i}_{\in \state}, \underbrace{\subgoal_i}_{\in \goal})
\end{align}
Figure~\ref{fig:architectures} (left) visualizes how the input is arranged and processed by Deep Sets.

In addition to permutation invariance, this design also gracefully handles input sets of arbitrary size, since the aggregation operation (summation) can pool together any number of elements. In our setting, this enables a single policy to handle environments of varying complexity and subtasks.

\begin{figure*}
    \centering
    \includegraphics[width=0.8\textwidth]{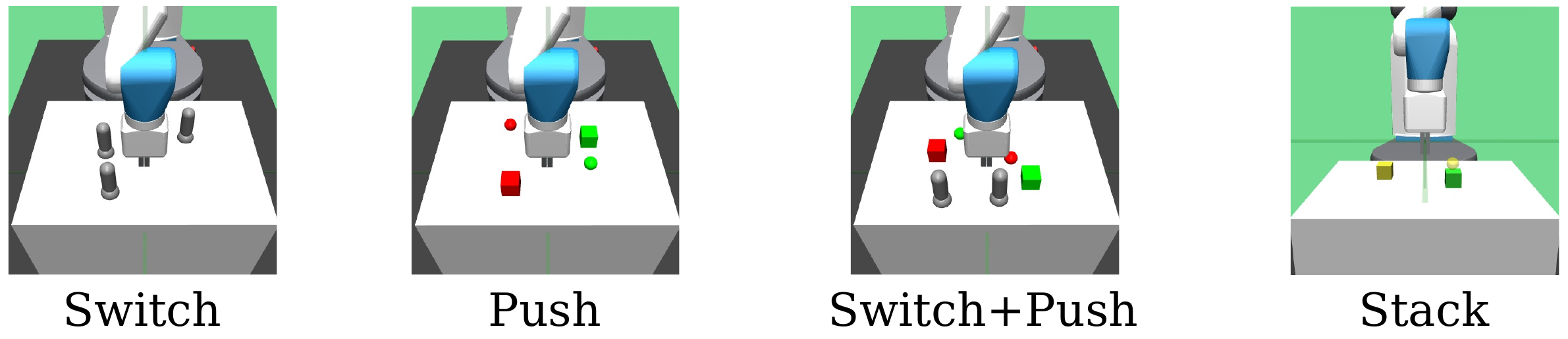}
    \caption{Illustrations of the robot manipulation environments we study. They consist of subtasks such as pushing a cube to its (spherical) target, flipping a switch to a specified position, and stacking one cube on top of another. The overall task can involve multiple entities and subtasks as well as their combinations (pushing cubes and flipping switches).}
    \label{fig:task_teaser}
\end{figure*}

\subsection{Self Attention}
Although the Deep Set policy satisfies permutation invariance (Proposition~\ref{prop:inv}) and is expressive enough to represent any $\pistar$ in theory, its design aggregates representations for all the  entities through a single summation and then lets the MLP $\rho$ handle any interactions between them. In tasks involving complex entity-entity interaction, we might desire a stronger inductive bias for modeling entity relationships.

\citet{transformer} introduced self-attention, a stackable mechanism for iteratively aggregating information across a sequence of vectors according to the relationships between them. Self attention linearly maps inputs $(x_1, \cdots, x_N)$ to three distinct sets of vectors: queries $(q_i)_{i=1}^N$, keys $(k_i)_{i=1}^N$, and values $(v_i)_{i=1}^N$. Self-attention then produces a sequence $(h_i)_{i=1}^N$ with the same dimensions as the input:
\begin{equation}
    \label{eq:self-attention}
    h_i = \sum_{j=1}^N \text{softmax}\left\{(q_i \cdot k_l)_{l=1}^N\right\}_j v_j
\end{equation}
To integrate this mechanism into a policy, we re-arrange the input $\state$ as a sequence $y=(y_1, \cdots, y_N)$, where vector $y_i$ contains all information relevant to entity $i$ (Eq.~\ref{eq:deepset-input}). The policy processes this sequence using an encoder architecture described in \citet{transformer}, which combines the self-attention mechanism with elementwise MLPs and normalization layers. We do \textbf{not} use causal masking or positional encodings, as we wish to preserve self-attention's innate permutation symmetry. The entire encoder $\text{SA}(\cdot)$ maps $y$ to another sequence $z=(z_1, \cdots, z_N)$ with identical dimensions while modeling relationships between the sequence elements. 

Since the policy must produce a single vector as output, we pool the $z_i$'s together by summation and project the result to an action $\action\in\actions$ using a small MLP $\rho(\cdot)$. The entire self attention policy's design is illustrated in Figure~\ref{fig:architectures} (right). Without the positional encodings, the output is invariant to permutations of the entities and subgoals, much like the Deep Set policy. However, the self-attention mechanism produces intermediate representations $z$ that includes interactions between the inputs, unlike Deep Set's independent intermediate representations.

\section{Experiments and Evaluation}
\label{sec:experiments}

In this section, we aim to study the following questions through our experimental evaluation.

\vspace*{-5pt}
\begin{enumerate}[leftmargin=*]
    \itemsep0em
    \item Can we \textbf{learn} more efficiently by using policies that utilize the structures and invariances of the EFMDP?
    \item Can the structured policies generalize better and enable \textbf{extrapolation} to more or fewer entities? 
    \item Can the structured policies solve tasks containing novel \textit{combinations} of subtasks, by \textbf{stitching} together (i.e. composing) learned skills?
\end{enumerate}
\vspace*{-5pt}

Extrapolation and stitching are particularly interesting as they require generalization to novel tasks with no additional training. This is particularly useful when deploying agents in real world settings with enormous task diversity.

\paragraph{Environment Description}
We seek to answer our experimental questions in a suite of simulated robotic manipulation environments, where the policy provides low level continuous actions to control a Fetch robot and interact with any number of cubes and switches. There are three subtasks: to \textit{push} a cube to a desired location on the table, to flip a \textit{switch} to a specified setting, or to \textit{stack} one cube on top of another. The higher level tasks can involve multiple cubes or switches and compose many subtasks together, as shown in Figure~\ref{fig:task_teaser}. These environments fit naturally into the EFMDP framework: the robot is the agent, the cubes and switches are entities, and the goal specifies desired cube locations or switch settings. 

We organize the environments into \textbf{families} to test learning and generalization. Environments in the \textit{N-Push} family require re-arranging $N$ cubes by pushing each one to its corresponding subgoal. The \textit{N-Switch} family requires flipping each of $N$ switches to its specified setting, and the \fnsnp{N} family involves re-arranging $N$ cubes \textit{and} flipping $N$ switches. We test extrapolation by varying $N$ within a family at test time, which changes the number of entities: for example we train a policy in \textit{3-Switch} and evaluate it in \textit{6-Switch}. As another example, we test 
stitching by training a single policy on \textit{2-Switch} and \textit{2-Push}, then evaluate it on \fnsnp{2} which requires combining the switch and pushing skills together in a single trajectory. Appendix~\ref{sec:envs} gives a full description of our environments.

\textbf{Baselines and Comparisons.}
Our main comparisons are with: (a) a baseline MLP that models the task as a regular MDP (Sec.~\ref{sec:mlp}), and (b) an ``oracle'' that manually coordinates solving one subtask at a time.

We construct subpolicies for the oracle by training one policy on each distinct subtask (pushing, flipping switches, and stacking). The oracle chooses an initial entity and subgoal arbitrarily, and uses the corresponding subpolicy until that subtask is solved. The oracle then selects the appropriate subpolicy for the next entity-subgoal pair and continues until the entire task is complete. The oracle is \textbf{not} guaranteed to achieve a $100\%$ success rate since it does not consider entity-entity interactions. An example failure mode is pushing one cube into position but knocking another one off the table while doing so. Still, as the oracle represents a hand-crafted hierarchical approach using an entity-based task decomposition, we will compare the RL and IL agents' performance against the oracle in the following experiments.

\subsection{Efficiency of Learning}

\begin{figure*}
    \centering
    \includegraphics[width=0.95\textwidth]{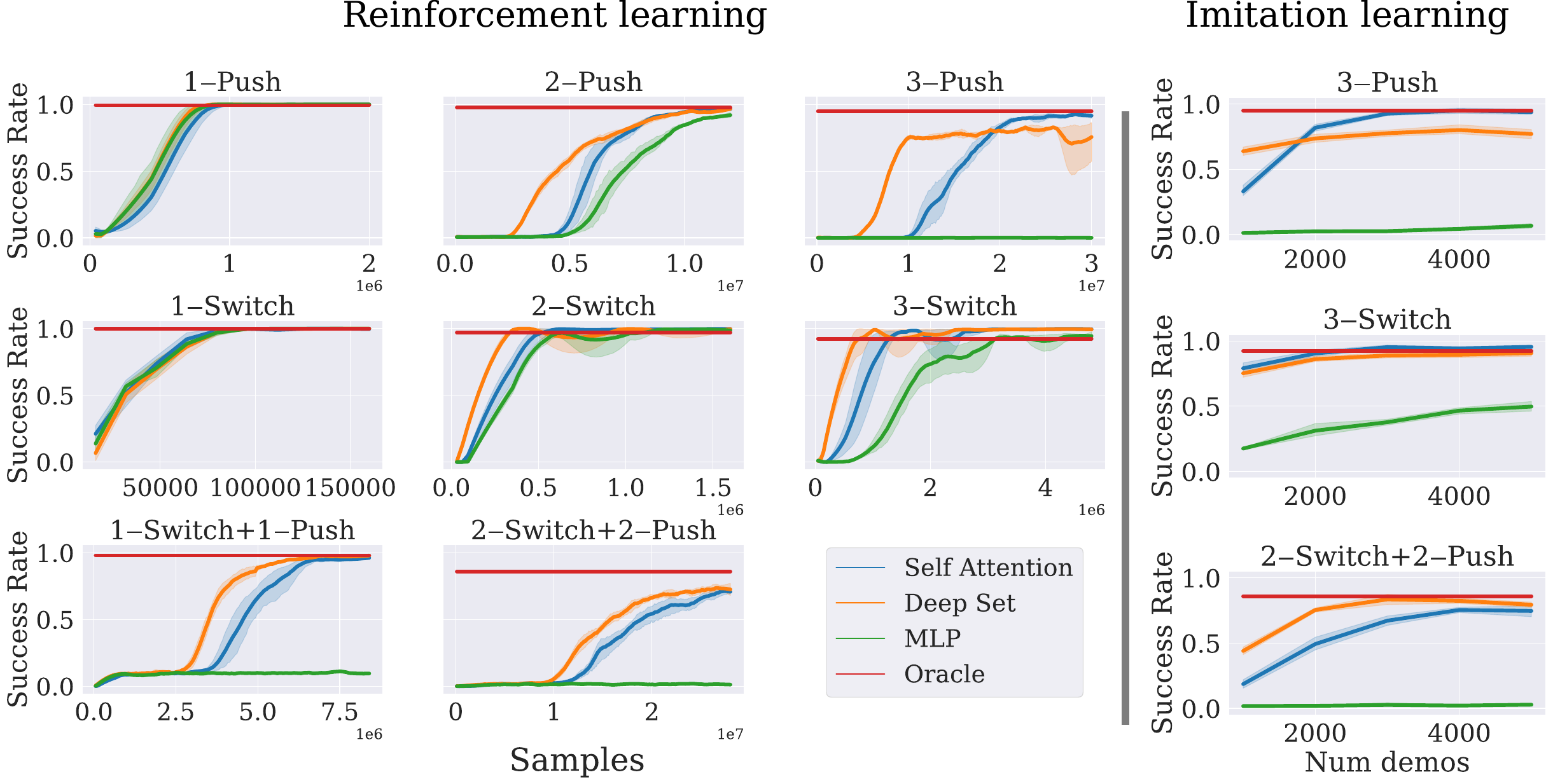}
    \caption{
    Training on environments of varying complexity using either reinforcement or imitation learning. Each row corresponds to a single environment family (\textit{N-Push}, \textit{N-Switch}, and \fnsnp{N}), where environments with larger $N$ contain more entities and are more complex. For RL (left), each plot is a training curve of success rate vs the number of steps taken in the environment. RL with standard MLPs can solve the simpler tasks such as \textit{1-Push}, but Deep Set and Self Attention policies are superior on the more complex environments. For IL (right), we show success rates of behavior cloning against number of expert demonstrations in the dataset. Both Deep Set and Self Attention policies far outperform the MLP even when given less data. Shaded regions indicate $95\%$ CIs over $5$ seeds.}
    \label{fig:learning}
\end{figure*}

To evaluate the learning efficiency of different architectures, we consider the \textit{N-Switch}, \textit{N-Push}, and  \fnsnp{N} environment families. We try $N=1, 2, 3$ for the first two families and $N=1, 2$ for the latter, with larger $N$ corresponding to more entities and more complex tasks within a family. \textbf{Evaluation criteria:} An episode in the environment is considered successful only if all the sub-goals in the environment are achieved.

We separately train policies on each environment in each family, using either RL or IL approaches. For RL training we use DDPG \citep{ddpg} with Hindsight Experience Replay (HER) \citep{her}, where we use the same architecture (either MLP, Deep Set, or Self Attention) to implement \textbf{both} the policy and critic. For IL we use behavior cloning to train policies to fit a dataset of expert trajectories using mean-squared error loss. For each environment, we use a trained RL agent to generate the corresponding expert trajectory datasets. See Appendix~\ref{sec:training} for full RL and IL training details.

\paragraph{RL results} Figure~\ref{fig:learning} (left) shows RL training curves as a function of environment samples.
In the simpler \textit{1-Switch} and \textit{1-Push} environments, all methods learn to solve the task fairly quickly. Once there is more than one entity, however, the Deep Set and Self Attention policies learn faster than the MLP. In harder environments like \textit{3-Push} or \fnsnp{N}, only the Deep Set and Self Attention policies achieve non-trivial success rate; these architectures match or exceed Oracle performance in all environments except \fnsnp{2}. Although they achieve similar asymptotic performance on most tasks, the Deep Set policy tends to learns faster than the Self Attention policy, possibly because the former is simpler and has fewer parameters.

\paragraph{IL Results} The imitation learning results appear in Figure~\ref{fig:learning} (right), where the x-axis now indicates the size of the training dataset used for behavior cloning. Similar to the RL setting, we see that the Deep Set and Self Attention policies learn far more efficiently than the MLP in all environments. For example, in \textit{3-Push} with $5000$ demonstrations, the MLP's success rate is still nearly zero while the Self Attention policy has a nearly $100\%$ success rate.

\paragraph{Conclusions} MLP policies struggle to learn complex tasks with many entities with both RL and IL, likely due to the lack of entity-centric processing that the Deep Set and Self Attention policies employ. The Deep Set policy typically learns faster than the Self Attention policy in RL, and matches or outperforms the transformer in IL with $1000$ trajectories. Although the asymptotic performance of the two entity-centric methods is typically similar, the Self Attention policy is better on \textit{3-Push} for both RL and IL. \textit{3-Push} is one of the more difficult tasks, and the Self Attention may benefit from greater relational expressivity through its self attention mechanisms. Overall, this experiment suggests that architectures that utilize the structure and invariances in EFMDPs learn substantially faster than black box architectures like the MLP.

\begin{figure*}[t]
    \centering
    \includegraphics[width=0.9\textwidth]{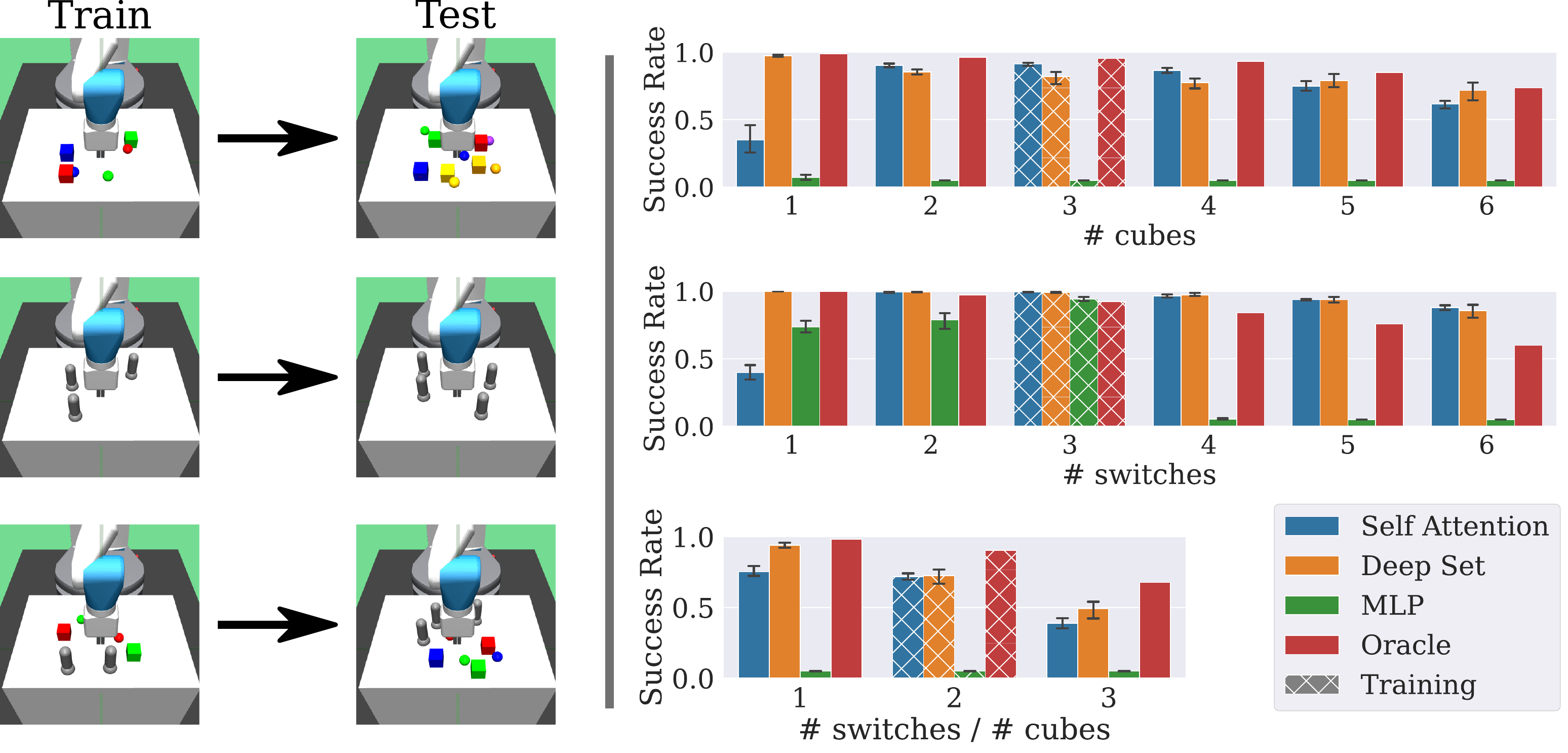}
    \caption{Extrapolation capabilities of RL-trained policies with different architectures. Each row is an environment family that contains environments with a varying number of entities. Policies are trained on a \textit{single} environment from each family before being tested on all the others, with no additional training. Bar charts show success rates in each environment, with the hatched bars corresponding to training environments. The Self Attention and Deep Set policies extrapolate beyond the training environment to solve tasks with more or fewer entities than seen in training, while MLP policies struggle on more complex testing environments. Error bars are $95\%$ CIs on 5 seeds.}
    \label{fig:extrapolation}
\end{figure*}

\subsection{Zero-Shot Extrapolation Capabilities}

To test whether trained policies can extrapolate and solve test tasks containing more or fewer entities than seen in training, we again use the \textit{N-Switch}, \textit{N-Push}, and \fnsnp{N} environment families. For the \textit{N-Push} and \textit{N-Switch} families we train a policy with RL on $N=3$ and test with $N\in\{1,\ldots,6\}$. For the \fnsnp{N} family we train a policy with RL on $N=2$ and test on $N \in\{1, \ldots, 3\}$. For testing, we use the RL agent checkpoint with the highest success rate in its training environment.

\textbf{Results and Observations} Figure~\ref{fig:extrapolation} shows the test performance of these policies on each environment family as the number of entities $N$ varies. The MLP only successfully learns the training task in the \textit{N-Switch} environments, and it generalizes decently to \textit{fewer} than $3$ switches, but fails completely in environments with \textit{more} than $3$ switches.

In contrast, the Deep Set and Self Attention policies generalize well and achieve zero-shot success rates comparable to or exceeding the Oracle in most test environments. Notably, these policies well exceed oracle performance on \textit{6-Switch} despite training in \textit{3-Switch}. Interestingly, Self Attention policies generalize poorly to test environments with only a single entity, such as \textit{1-Push} or \textit{1-Switch}. This suggests that the self attention mechanism may rely critically on interactions between entities during training, which generalizes poorly to the single entity case. The Deep Set policy extrapolates better than the Self Attention policy on most environments. Overall, we find that by using architectures capable of utilizing the EFMDP structure, agents can perform very effective extrapolation.

\subsection{Zero-Shot Stitching to solve novel tasks}
\begin{figure*}
\includegraphics[width=\textwidth]{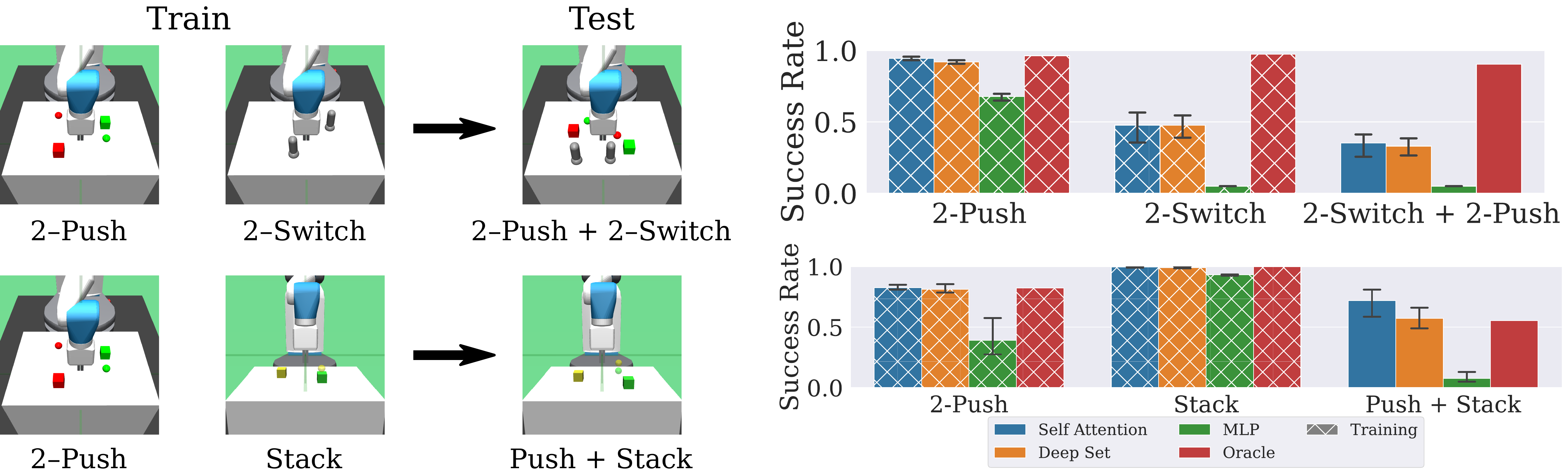}
\caption{Left: each row corresponds to a setting where policies are evaluated on test tasks that require stitching together skills learned in training, with no additional data. In the top setting, a single agent is trained on \textit{2-Switch} and \textit{2-Push}, and must solve \textit{2-Switch+2-Push} at test time. In the bottom setting, a single agent is trained on \textit{2-Push} and \textit{Stack}, and must solve \textit{Push + Stack} at test time. Right: average success rates of policies with different architectures in each setting. Deep Set and Self Attention policies are moderate successful at solving the test tasks, and are comparable to the Oracle in \textit{Push + Stack}. The MLP fails to achieve nontrivial success rates on both test environments. Error bars indicate $95\%$ CIs over $5$ seeds.
}
\label{fig:stitching}
\end{figure*}
When evaluating policies for stitching behavior, we use test tasks that combine subtasks from training in novel ways. In our first setting, we train a policy on \textit{2-Push} and \textit{2-Switch}, and then test this policy on \fnsnp{2}, which requires both pushing cubes \textit{and} flipping switches. In our second setting, we train a single policy on \textit{2-Push} and \textit{Stack}, which requires stacking one cube on top of another. The test environment is \textit{Push + Stack}, which requires pushing one cube into position and then stacking the other block on top. This setting is especially difficult because it requires zero-shot stitching of skills in a particular order (push, \textit{then} stack). Figure~\ref{fig:stitching} (left) shows the train-test task relationships we use to test stitching.

\textbf{Results and Observations} \ Since this experiment requires training a single policy on multiple training tasks, during each episode we choose one of the training tasks uniformly at random. Figure~\ref{fig:stitching} (right) shows that the MLP policy fails to jointly learn the training tasks in the first setting, leading to poor performance in \fnsnp{2}. However, the MLP averages above a $35\%$ success rate on both training tasks in the second setting, but still only manages a $5\%$ success rate on \textit{Push + Stack}. This suggests that even when MLP policies are capable of learning the training tasks, they are unable to combine them to solve new ones.

The Deep Set and Self Attention architectures show substantially better stitching capabilities compared to the MLP, but are not as competent as the oracle. It is particularly surprising that the Self Attention policy achieves a $60\%$ zero-shot success rate on \textit{Push + Stack}, which requires understanding that the push and stack subtasks must be executed in a specific order. Poor performance in \fnsnp{2} is again likely due to difficulties in training one policy on two different tasks. For example, the Deep Set policy has a $100\%$ success rate when learning \textit{2-Switch} alone (Figure~\ref{fig:learning}), but only a $48\%$ success rate here when also jointly learning \textit{2-Push}. This suggests that improving joint training could further improve stitching performance.

\section{Conclusion}
In this work, we introduced the EFMDP framework for the learning paradigm where an agent can interact with many entities in an environment. We explore the structural properties of this framework like invariance of the reward and dynamics under permutation symmetry. Using this structure, we showed that the optimal policy and value function in the EFMDP are also invariant to permutations of the entities.

Building on the above result, we introduced policy architectures based on Self-Attention and Deep Sets that can leverage the symmetries and invariances in the EFMDP. More specifically, these policy architectures decompose goal-conditioned tasks into their constituent entities and subgoals. These architectures are flexible, do not require any manual task annotations or action primitives, and can be trained end-to-end with standard RL and IL algorithms.

Experimentally, we find that our policy architectures that utilize the EFMDP structure can: (a) \textbf{learn substantially faster} than black-box architectures like the MLP; (b) perform \textbf{zero-shot extrapolation} to new environments with more of fewer entities than observed in training; and (c) perform \textbf{zero-shot stitching} of learned behaviors to solve novel task combinations never seen in training.

\textbf{Limitations and Future Work: } As a first step in this line of work, we focused on the case where the agent and entity state spaces are compact vectors. In future work, we hope to extend these ideas to high-dimensional observation spaces like visual inputs. Learning object-centric representations from high-dimensional inputs is a vibrant research area~\citep{burgess2019monet,StructuredWorldModel,SlotAttention,nanbo2020learning}, and such techniques could help extend our results to visual control policies that learn efficiently and generalize well.

\section*{Acknowledgements}
We would like to thank Kevin Lu, Amy Zhang, Eugene Vinitsky, and Karl Pertsh for helpful discussions and advice throughout this paper's development. We also thank Abhinav Gupta for valuable feedback and mentorship, and Shagun Sodhani for code reviews.

\FloatBarrier
\bibliography{example_paper}
\bibliographystyle{icml2022}

\newpage
\appendix
\onecolumn
\section{Permutation invariance}
\label{sec:permutation-invariance}
We recall Proposition~\ref{prop:inv}:

\inv*

We want to show that any optimal policy $\pistar:\states\times\goals\rightarrow\actions$ and the optimal action-value function $\qstar:\states\times\actions\times\goals\rightarrow\reals$ are both permutation invariant, that is for any $\sigma\in S_N$:
\begin{align}
\pistar(\sigma \state, \sigma \goal) = \pistar(\state, \goal)\\
\qstar(\sigma \state, \action, \sigma \goal) = \qstar(\state,\action,\goal)
\end{align}
Recall that in an EFMDP the reward and dynamics have permutation symmetry (Property~\ref{property:inv}):
\begin{align*}
    \rewards(\state, \action, \goal) = \rewards(\sigma\state,\action,\sigma\goal)\\
    \transition(\state'|\state,\action) = \transition(\sigma \state'|\sigma \state, \action)
\end{align*}
where $\sigma\state$ and $\sigma\goal$ are defined in Eq.~\ref{eq:sigma-s} and Eq.~\ref{eq:sigma-g}. We assume for simplicity that the agent space $\agents$ and entity space $\entities$ are discrete, so that the state space $\states=\agents\times\entities^N$ is also discrete.

We begin with $\qstar$, which can be obtained by value iteration, where $\qstar_k$ denotes the $k$'th iterate. We initialize $\qstar_0 \equiv 0$, which is (trivially) permutation invariant. Permutation invariance is then preserved during each step of value iteration $\qstar_k \mapsto \qstar_{k+1}$:
\begin{align}
    \qstar_{k+1}(\sigma\state, \action, \sigma\goal) &= \rewards(\sigma\state, \action, \sigma\goal) + \gamma\max_{\action'} \sum_{\state' \in \states} \transition(\state'|\sigma \state, \action)\qstar_k(\state',\action')\\
    \label{eq:proof-s1}
    &= \rewards(\state, \action, \goal) + \gamma\max_{\action'} \sum_{\state'\in\states} \transition(\sigma^{-1}\state'|\state, \action)\qstar_k(\sigma^{-1}\state',\action')\\
    \label{eq:proof-s2}
    &= \rewards(\state, \action, \goal) + \gamma\max_{\action'} \sum_{\state' \in \states} \transition(\state'|\state, \action)\qstar_k(\state',\action')\\
    &= \qstar_{k+1}(\state, \action,  \goal)
\end{align}
Hence $\qstar_k$ is permutation invariant for all $k=0, 1, \cdots$, with $\qstar_k\xrightarrow[k\rightarrow\infty]{}\qstar$. Line~\ref{eq:proof-s1} follows from the permutation invariance of the reward, transition probability, and the previous iterate $\qstar_k$. Line~\ref{eq:proof-s2} uses the fact that summing over $\sigma^{-1}\state'$ for all $\state' \in \states$ is the same as simply summing over all states $\state'\in\states$.
This can be seen more explicitly by expanding a sum over arbitrary function $f(\cdot)$:
\begin{equation*}
    \sum_{\state\in\states} f(\sigma^{-1}\state) = \sum_{\agent\in\agents}\sum_{\entity_1\in\entities}\cdots\sum_{\entity_N\in\entities} f(\agent, \entity_{\sigma^{-1}(1)}, \cdots, \entity_{\sigma^{-1}(N)}) = \sum_{\state \in \states} f(\state)
\end{equation*}

The permutation invariance of $\qstar$ leads to the permutation invariance of $\pistar$:
\begin{equation*}
    \pistar(\sigma \state, \sigma \goal) = \arg\max_\action \qstar(\sigma\state,\action, \sigma \goal) = \arg\max_\action \qstar(\state,\action,\goal) = \pistar(\state, \goal)
\end{equation*}

\section{Environments}
\label{sec:envs}
Our environments are modified from OpenAI Gym's Fetch environments~\citep{gym}, with our stacking environment in particular being modified from the Fetch stacking environments of \citet{lanier2019curiosity}. They have a 4D continuous action space with 3 values for end effector displacement and 1 value for controlling the distance between the gripper fingers. The final action is disabled when the neither the training or test tasks involve stacking, since gripping is not required for block pushing or switch flipping. Input actions are scaled and bounded to be between $[-1, 1]$.
We set the environment episode length based on the number of entities and subtasks involved. Each switch added $20$ timesteps, and each cube pushing or stacking task added $50$ timesteps. For example, \fnsnp{2} had a max episode length of $2 \times 50 + 2 \times 20 = 140$ timesteps.

For non-stacking settings such as \textit{N-Push} and \textit{N-Switch + N-Push}, we disable cube-cube collision physics to make training easier for all methods. Note that subgoals may still interfere with each other since the gripper can interact with all cubes, so the agent may accidentally knock one cube away when manipulating another one. We repeat the extrapolation experiments for \textit{N-Push} \textit{with collisions} in Appendix~\ref{sec:collisions}.

\textbf{State and goals.} The agent state describe the robot's end effector position and velocity the gripper finger's positions and velocities. The entity state for cubes include the cube's pose and velocity, and for switches include the switch setting $\theta \in [-0.7, 0.7]$ and the position of the switch base on the table. The switch entity state is padded with zeros to match the shape of the cube entity state, and all entity states include an extra bit to distinguish cubes from switches. Subgoals specify a target position for cubes and a target setting $\theta^\star\in\{-0.7, 0.7\}$ for switches.

\textbf{Reward.} The dense reward is defined as the average distance between each entity and its desired state as specified by the subgoal. For cubes, this is the L2 distance between current and desired position. For switches, this is $|\theta-\theta^\star|$, where $\theta$ is the current angle of the switch and $\theta^\star$ is the desired setting. The sparse reward is $0$ if all entities are within a threshold distance of their subgoals, and $-1$ otherwise.

\section{Training details}
\label{sec:training}
\subsection{Reinforcement learning}
We train RL agents using a publicly available implementation\footnote{\url{https://github.com/TianhongDai/hindsight-experience-replay}} of DDPG~\citep{ddpg} and Hindsight Experience Replay (HER)~\citep{her}. Table~\ref{table:rl-hparam} contains the default hyperparameters shared across all experiments. Our modified implementation collects experience from $16$ environments in parallel into a single replay buffer, and trains the policy and critic networks on a single GPU. We collect $2$ episodes for every $5$ gradient updates, and for HER we relabel the goals in $80\%$ of sampled minibatches (the ``relabel prob''). The reward scale is simply a multipler of the collected reward used during DDPG training. For exploration we use action noise $\eta$ and random action probability $\epsilon$; the output action is:
    \begin{equation*}
    \tilde{a} \sim
    \begin{cases}
        a + \mathcal{N}(0, \eta), &\text{with prob } 1-\epsilon\\
        \text{Uniform}(-1, 1), &\text{with prob } \epsilon
    \end{cases}
\end{equation*}

Table~\ref{table:env-rl-params} shows environment specific RL hyperparameters. ``Epochs'' describes the total amount of RL training done, with 1 epoch corresponding to $50\times\textsc{parallel envs}$ episodes. Sparse reward is used for the simpler environments, and dense reward for the harder ones. For some environments we decay the exploration parameters $\eta,\epsilon$ by a ratio computed per-epoch. Lin(.01, 100, 150) means that $\eta,\epsilon$ are both decayed linearly from $\eta_0$ and $\epsilon_0$ to $.01\times\eta_0$ and $.01\times\epsilon_0$ between epochs $100$ and $150$. The constant exploration decay schedule maintains the initial $\eta_0, \epsilon_0$ values throughout training. The target network parameters are updated as $\theta^{\text{target}} \gets (1-\tau)\theta + \tau\theta^{\text{target}}$, where $\tau$ is the target update speed.

We use the same RL hyperparameters regardless of architecture type except that the learning rate is lower for Self Attention and the exploration decay schedule may vary. Where Table~\ref{table:rl-hparam} lists ``Fast'' and ``Slow'' decay schedules, we sweep over both options for each architecture and use the schedule that works best. In each case, the Self Attention policy prefers the slower exploration schedule and Deep Sets prefers the faster one, while the MLP typically fails to learn with either exploration schedule on the more complex environments.

\textbf{Architectures.} The exact actor and critic architectures uses for each architecture family is shown in Table~\ref{table:rl-architectures}. Linear(256) represents an affine layer with $256$ output units. ReLU activations follow every layer except the last. The final actor layer is followed by a Tanh nonlinearity, and the critic has no activation function after the final layer. $A$ represents the action space dimension, and Block(N, M, H) represents a Transformer encoder block~\citep{transformer} with embedding size $N$, feedforward dimension $M$, and $H$ heads. We disable dropout within the Transformer blocks for RL training.

\begin{table}
\caption{General shared RL hyperparameters}
\label{table:rl-hparam}
\begin{center}
\begin{tabular}{||c | c||} 
 \hline
 Hyperparameter & Value \\ [0.5ex] 
 \hline\hline
 Discount $\gamma$ & 0.98 \\
 \hline
 Parallel envs & 16 \\
 \hline
 Replay buffer size & $10^6$\\
 \hline
 Relabel prob & 0.8 \\
 \hline
 Ratio of episodes : updates & $2:5$\\
 \hline
 Optimizer & Adam \\
 \hline
 Learning rate & MLP, Deep Set: 0.001 \\
 & Self Attention: 0.0001 \\
 \hline
 Reward Scale & Sparse: 1; Dense: 5\\
 \hline
 Action noise $\eta_0$ (initial) & 0.2 \\
 \hline
 Random action prob $\epsilon_0$ (initial) & 0.3 \\
 \hline
\end{tabular}
\end{center}
\end{table}
\begin{table}
\caption{Environment specific RL hyperparameters}
\label{table:env-rl-params}
\begin{center}
\begin{tabular}{||c | c | c | c | c||} 
 \hline
 Environment & Reward & Epochs & Exploration decay & Target update speed $\tau$\\ [0.5ex] 
 \hline\hline
 1-Push & Sparse & 50 & Constant(1) & 0.95\\
 \hline
 2-Push & Dense & 150 & Lin(.01, 75, 125) & 0.99\\
 \hline
 3-Push & Dense & 250 & Fast: Lin(.01, 30, 80) & 0.99\\
 & & & Slow: Lin(.01, 100, 175) & \\
 \hline
 \{1,2,3\}-Switch & Sparse & \{10, 50, 100\} & Constant(1) & 0.95\\
 \hline
 1-Switch + 1-Push & Dense & 150 & Lin(.01, 60, 100) & 0.99\\
 \hline
 2-Switch + 2-Push & Dense & 250 & Fast: Lin(.01, 75, 150) & 0.99\\
 & & & Slow: Lin(.01, 100, 150) & \\
 \hline
\end{tabular}
\end{center}
\end{table}
\begin{table}
\caption{RL architectures}
\label{table:rl-architectures}
\begin{center}
\begin{tabular}{||c | c | c||} 
 \hline
 Family & Actor & Critic \\ [0.5ex] 
 \hline\hline
 MLP & Linear(256)$\times 3$, Linear(A) & Linear(256) $\times 3$, Linear(1)\\
 \hline
 Deep Set & $\phi$: Linear(256) $\times 3$ & $\phi$: Linear(256) $\times 2$\\
 & $\rho$: Linear(A) & $\rho$: Linear(256), Linear(1) \\
 \hline
 Self Attention & SA: Linear(256), Block(256, 256, 4)$\times 2$ & SA: Linear(256), Block(256, 256, 4)$\times 2$\\
 & $\rho$: Linear(A) & $\rho$: Linear(A)\\
 \hline
\end{tabular}
\end{center}
\end{table}
\subsection{Imitation Learning}
The IL dataset is generated using the best performing RL agent in that environment--we record $M\in\{1000, 2000, 3000, 4000, 5000\}$ demonstration trajectories. This creates a dataset of $M \times T$ transitions $\mathcal{D} = \{(\state_i,\action_i)\}_{i=1}^{M\times T}$ for behavior cloning. However, in practice we filter the dataset slightly by discarding the transitions corresponding to trajectories that are not successful.

We use the same policy architectures shown in Table~\ref{table:rl-architectures} and optimize mean squared error loss over the dataset:
\begin{equation*}
    \arg\min_\pi J(\pi) \defeq \frac{1}{|\mathcal{D}|}\sum_{(\state, \action) \sim \mathcal{D}} ||\pi(\state) - \action||^2
\end{equation*}
We use the Adam~\citep{adam} optimizer with learning rate $0.001$ (MLP, Deep Sets) or $0.0001$ (Self Attention). Each policy is trained for $60,000$ gradient steps with a batch size of $128$.

\subsection{Training and inference speed}
Here we consider the computational complexity of using different architecture classes (MLPs, Deep Sets, and Self Attention), as we scale the number of entities $N$. We consider the number of parameters, activation memory, and computation time (for a forward pass). For MLPs with fixed hidden layer sizes, the number of parameters and computation time increase linearly with $N$ while the memory required for activations stays fixed (due to fixed hidden layer sizes). In Deep Sets and Self Attention, the number of parameters does not depend on the number of entities $N$. The activation memory and computation time grow linearly in Deep Sets, and quadratically for the pairwise interactions of Self Attention. In practice, the number of entities $N$ is modest in all our environments (e.g., fewer than $10$), but computational complexity may be relevant in more complex scenes with lots of entities.

\begin{figure}[!htb]
    \centering
    \includegraphics[width=\textwidth]{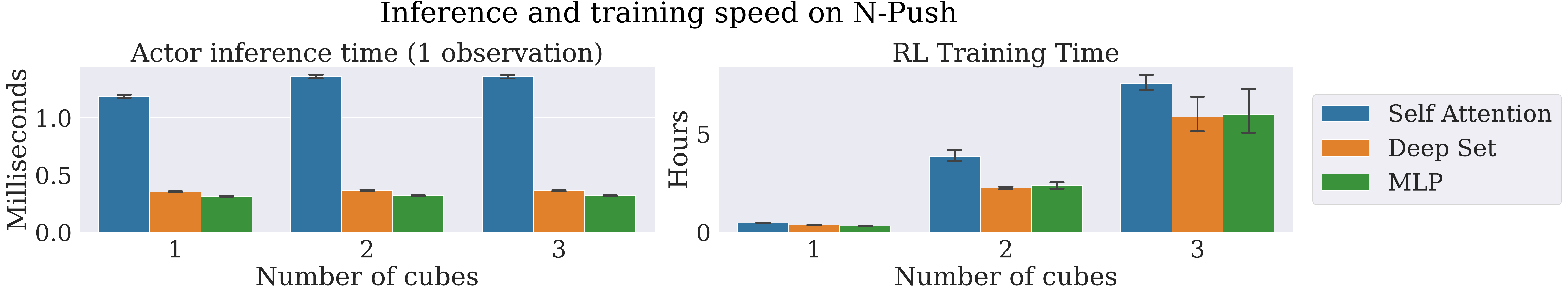}
    \caption{Left: the time (in milliseconds) it takes for each policy architecture to execute a single forward pass on a single input observation from the \textit{N-Push} environments, where $N\in\{1, 2, 3\}$. The self attention policy is significantly slower, while the Deep Set and MLP policies are comparable. Right: Real world reinforcement learning times (in hours) training each policy/critic architecture on the \textit{N-Push} environments. Although the Self Attention policy is slightly slower, all policies train at comparable speeds in the same environment. This suggests that environment simulation, not policy execution, is the dominant time consuming element.}
    \label{fig:rl-trainspeed}
\end{figure}

For a more holistic real-world comparison of execution and training speed, Figure~\ref{fig:rl-trainspeed} shows both inference time and training time in the \textit{N-Push} environments for $N\in\{1, 2, 3\}$. The inference time is the number of milliseconds it takes an actor do a single forward pass (using a GPU) on a single input observation. The Self Attention policy involves more complex computations and is significantly slower than Deep Set and MLP policies. The RL training time is the actual number of hours required to run the reinforcement learning algorithms of Figure~\ref{fig:learning}, for each architecture. Not surprisingly, we see that \textit{3-Push} takes significantly longer to train than \textit{1-Push}, since it is a harder environment. For a fixed environment, however, all three architecture types are comparable in speed, with the Self Attention version being slightly slower than the others. The surprising similarity in RL training time (despite much slower inference time for the Self Attention policy) suggests that most of the RL time is spent on environment simulation rather than policy or critic execution. Hence, the difference between architectures presented in this paper has only a minor effect on reinforcement learning speeds in practice.

\section{Further comparisons}
\subsection{Deep set architecture size}
\begin{figure}[!htb]
    \centering
    \includegraphics[width=\textwidth]{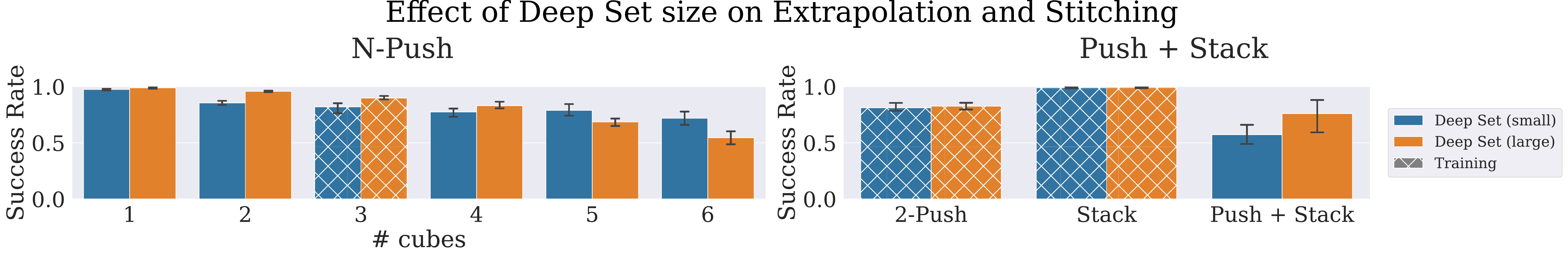}
    \vspace{-2em}
    \caption{Comparison of \textit{N-Push} extrapolation and \textit{Push + Stack} stitching performance when using small and large variants of the Deep Set policy architecture. The small version implements $\rho$ with a 1-layer linear map, while the large version implements $\rho$ with a 2-layer MLP. For \textit{N-Push}, the larger network achieves greater success rates in the training environment (3 cubes) but is actually worse when extrapolating to 5 or 6 cubes. On the other hand, the larger Deep Set displays superior stitching capability and achieves a higher average success rate when generalizing to \textit{Push + Stack} from \textit{2-Push} and \textit{Stack}.}
    \label{fig:deepset_size}
\end{figure}

Recall that our Deep Set policy architecture involves two MLPs $\phi$ and $\rho$, where $\phi$ produces intermediate representations for each entity, those intermediate representations are summed, and then $\rho$ produces the final output (Eq.~\ref{eq:deep-sets}). In full generality, both $\phi$ and $\rho$ may have two or more layers with nonlinearities in between. While our $\phi$ is a 3-layer MLP, we use a \textit{linear} $\rho$ throughout the main paper because we found that it often works comparably or better than using a larger 2-layer MLP $\rho$. Here we repeat the \textit{N-Push} extrapolation and \textit{Push + Stack} stitching experiments from the main paper using a 2-layer $\rho$, which we call ``Deep Set (large).'' The results from the main paper uses a 1-layer $\rho$ which we refer to here as ``Deep Set (small).''

Figure~\ref{fig:deepset_size} shows the results. In \textit{N-Push}, the larger Deep Set model achieves higher training success rates in the 3-cube environment, but has worse extrapolation success rates for large numbers of cubes. For example, the smaller Deep Set model is significantly better at solving \textit{6-Push}. Meanwhile, the large and small Deep Sets achieve very similar results in the pushing and stacking training environments. However, the larger Deep Set model achieves a higher success rate in the \textit{Push + Stack} environment, indicating superior stitching capability. This suggests that simpler Deep Set architectures may be better for extrapolating to a large number of entities, but more complex architectures may be superior for solving complex tasks with a fixed number of entities.

\subsection{N-Push with cube-cube collisions}
\label{sec:collisions}
\begin{figure}[!htb]
    \centering
    \includegraphics[width=0.7\textwidth]{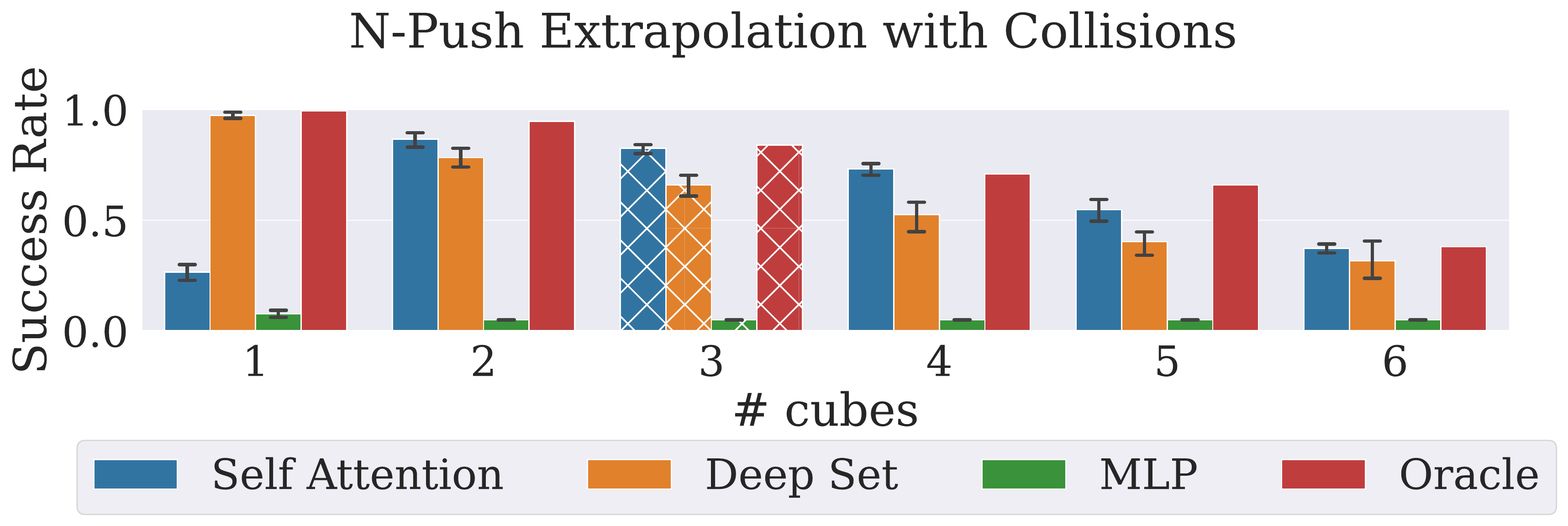}
    \caption{\textit{N-Push} extrapolation with cube-cube collisions enabled. All methods observe some drop in performance relative to Figure~\ref{fig:extrapolation}, where \textit{N-Push} has cube-cube collisions disabled. Self Attention tends to outperform Deep Sets when collisions enabled, likely because its relational inductive biases are better suited to handling interactions between entities that arise from collisions.}
    \label{fig:cube_collisions}
\end{figure}
As noted in Appendix~\ref{sec:envs}, we disable cube-cube collisions in the \textit{N-Push} and \textit{N-Switch+N-Push} experiments of the main paper (of course, the stacking settings require cube-cube collisions to be enabled). Here we repeat the \textit{N-Push} extrapolation experiments with cube-cube collisions \textit{enabled}. Figure~\ref{fig:cube_collisions} shows the results, which are qualitatively similar to when collisions are disabled. All methods observe a decrease in success rates of about $15\%$, with the Self Attention method often outperforming the Deep Set policy. This is likely because \textit{N-Push} involves more interaction between entities once cube-cube collisions are enabled, and Self Attention's relational inductive biases are better suited for modeling these interactions.

\end{document}